\documentclass[runningheads]{llncs}

\usepackage{eccv}

\usepackage{float}
\usepackage{placeins}
\usepackage{eccvabbrv}

\usepackage{graphicx}
\usepackage{booktabs}

\usepackage[accsupp]{axessibility}  %
\usepackage{multirow}

\usepackage{hyperref}

\usepackage{orcidlink}

\begin{document}

\title{ Ig3D: Integrating 3D Face Representations in Facial Expression Inference} 

\titlerunning{Ig3D}

\author{Lu Dong $^{*}$ \orcidlink{0009-0007-4036-7690} \and
Xiao Wang $^{*}$ \orcidlink{0009-0000-0511-8495} \and
Srirangaraj Setlur\orcidlink{0000-0002-7118-9280}
\and
Venu Govindaraju\orcidlink{0000-0002-5318-7409}
\and
Ifeoma Nwogu \orcidlink{0000-0003-1414-6433}}

\authorrunning{L.~Dong et al.}

\institute{State University of New York at Buffalo\\
\url{https://dongludeeplearning.github.io/Ig3D.html} }

\maketitle
\begingroup
\renewcommand\thefootnote{*}
\footnotetext{Equal contribution.}
\endgroup

\begin{abstract}
Reconstructing 3D faces with facial geometry from single images has allowed for major advances in animation, generative models, and virtual reality. However, this ability to represent faces with their 3D features is not as fully explored by the facial expression inference (FEI) community.
This study therefore aims to investigate the impacts of integrating such 3D representations into the FEI task, specifically for facial expression classification and face-based valence-arousal (VA) estimation.
To accomplish this, we first assess the performance of two 3D face representations (both based on the 3D morphable model, FLAME) for the FEI tasks. We further explore two fusion architectures, intermediate fusion and late fusion, for integrating the 3D face representations with existing 2D inference frameworks. To evaluate our proposed architecture, we extract the corresponding 3D representations and perform extensive tests on the AffectNet and RAF-DB datasets.  
Our experimental results demonstrate that our proposed method outperforms the state-of-the-art AffectNet VA estimation and RAF-DB classification tasks. Moreover, our method can act as a complement to other existing methods to boost performance in many emotion inference tasks. 

\end{abstract}

\section{Introduction}

Facial expressions play a significant role in social interactions, as it can provide insights into a person’s feelings toward other individuals or events. Mehrabian and Wiener \cite{mehrabian1967decoding} suggest that 55\% of communication is perceived through facial expressions. 
AI-based automated emotion analysis enhances user experience and it has well-known applications in autonomous driving \cite{chen2022emotion}, online course learning\cite{bian2019spontaneous}, security \cite{nan2022mobilenet,li2021facial}, healthcare\cite{munsif2024optimized}, medical rehabilitation \cite{yun2017social}, employee job retention\cite{prentice2020emotional}, and many other social situations.

In the domain of facial expression inference (FEI)\footnote{Wagner \etal \cite{wagner2024cage} postulate that facial expressions are inferred and not recognized, hence the use of the term FEI, rather than the more popular term FER - facial expression recognition.},
there are two primary tasks: discrete and continuous facial expression inference. Discrete or categorical inference aims to assign facial expressions into distinct emotional categories, such as anger, sadness, joy/happiness, surprise, fear, disgust, contempt and neutral, 
whereas continuous inference assigns facial expressions within a continuous 2D numerical space; where the two dimensions are extents of \emph{valence} (the extent of pleasant or unpleasant response to an emotional stimulus) and \emph{arousal} (the state of physiological activation and alertness resulting from an emotional stimulus).
Discrete classification has advanced quickly due to the ease of data annotation. However, the highly abstract nature of labels makes cross-cultural consensus difficult\cite{barrett2019emotional}. Additionally, experts disagree on the number of emotion categories \cite{epstein1984controversial,roseman1984cognitive, izard1992basic,ekman1992argument}, which have increased from 7 to 135 \cite{chen2022semantic}. Continuous circumplex emotion modeling \cite{russell1980circumplex} further quantifies emotions and promotes research, but it is challenging to annotate accurately and suffers from bias \cite{mollahosseini2017affectnet,kollias2021affect}. To mitigate this, researchers typically try to increase the number of human observers to reduce bias \cite{ren2024veatic}. Although both types of analysis have certain data limitations, combining them offers complementary benefits. Therefore, we believe that the model's analysis should consider both perspectives.

Recently, 3D mesh reconstruction from static human-centric images has shown impressive achievements across various applications \cite{zhai2023language,shan2024towards, dong2024signavatar, xu2024comparative}. In particular, reconstructing 3D faces from monocular images using facial geometry has proven effective in capturing extreme, asymmetric, and subtle expressions accurately.
Regressing parameters from images as a lightweight 3D representation can disentangle facial shape and expression, and easily generate 3D facial geometry using a morphable model like FLAME \cite{li2017learning}.
Therefore, a natural progression is to integrate this technology into facial expression inference. However, limited research has investigated the significance and impact of the parameters involved, and how they can enhance FEI. In this work, we investigate the performance of two latest 3D face regression models, EMOCA\cite{danvevcek2022emoca} and SMIRK\cite{smirk20243d}, in the context of the FEI task. 

Data fusion is a process dealing with data and information from multiple sources to achieve improved information for decision-making \cite{hall1997introduction}.
This paper addresses the fusion of information from 2D images with parameters regressed from 3D perspectives, presenting significant challenges and uncertainties. A major issue is the heterogeneity of feature representations. Features from different modalities can vary greatly, requiring effective methods to integrate these diverse features seamlessly.
Therefore, this paper proposes two architectures, intermediate fusion and late fusion, to investigate the impact of different fusion methods on facial expression inference performance.

\noindent In summary, the contributions of this work are as follows:

\begin{itemize}
    \item[] First, We provide insights into the parameters of 3D face representation (pose, shape, expression, jaw, etc.) and compare the two recent 3D face representation models, SMIRK and EMOCA on the benchmark datasets. Experiments showcase that EMOCA 3D representation achieves better performance on FEI tasks.
    \medskip

    \item[] We introduced two architectures for integrating 3D representations: intermediate fusion and late fusion. Our experiments demonstrate that the late fusion architecture achieves superior performance.
    \medskip

     \item[] Lastly, we present a simple and effective architecture that can be flexibly adapted to various affective reasoning tasks. Extensive experiments demonstrate the efficiency of our method, with results surpassing the state-of-the-art in AffectNet VA estimation and RAF-DB classification. We will release the 3D representation data along with the code developed for this work.
\end{itemize}

\section{Related Work}

\subsection*{Facial Expression Inference Datasets}
    The continuous development of datasets has driven the advancement of AI-based affective models. Early datasets such as JAFFE \cite{lyons1998japanese}, CK+ \cite{lucey2010extended}, and KDEF \cite{calvo2008facial} primarily collected expressions under 7 or 8 discrete categories. Subsequently, in-the-wild datasets like FER2013\cite{goodfellow2015challenges}, AffectNet \cite{mollahosseini2017affectnet}, RAF-DB\cite{li2017reliable,li2019reliable}, and Aff-Wild2 \cite{kollias20247th, kollias20246th, kollias2023abaw2, kollias2023multi, kollias2023abaw, kollias2022abaw, kollias2021analysing,kollias2021affect, kollias2021distribution, kollias2020analysing, kollias2019expression, kollias2019deep, kollias2019face, zafeiriou2017aff} further increased the scale of data.  EmotionNet \cite{benitez-quinones2020emotionet},  LIRIS\_CSE\cite{khan2019novel} and MAFW \cite{liu2022mafw} extended the discrete labels by including compound annotations to better capture the richness of human emotions. Additionally, fine-grained discrete labels have been explored in datasets like F2ED\cite{zhang2020fine} and Emo135 \cite{wang2021semantic}.  Having both discrete and continuous annotations also opens up new avenues for research breadth \cite{antoniadis2021audiovisual,mollahosseini2017affectnet,kosti2017emotion} Contextual information is consciously considered and collected in the EMOTIC\cite{kosti2017emotion} and VEATIC\cite{ren2024veatic} datasets. 

To better compare performance advantages, this paper selects the widely studied AffectNet and RAF-DB datasets for analysis. AffectNet includes 287,651 training images and 3,999 validation images (typically used as test data in experiments). It provides annotations for 8 distinct categories and VA (Valence-Arousal) annotations. 
RAF-DB includes 12,272 images for training and 3,096 images for testing, covering 7 distinct categories. Additionally, the dataset contains labels related to attributes such as gender, age, and more.

\subsection*{Facial Expression Inference Models}

Discrete expression inference on datasets like AffectNet and RAF-DB has made continuous progress. The current top accuracy models \cite{PaperwithCode} on both datasets are the same: DDAMFN, FMAE, BTN, and ARBEx. 
DDAMFN \cite{zhang2023dual} integrates a Mixed Feature Network (MFN) as the backbone and a Dual-Direction Attention Network (DDAN) as the head.
FMAE \cite{ning2024representation} introduces Identity Adversarial Training (IAT) and pre-trains a Facial Mask Autoencoder. S2D \cite{chen2023static} proposes the Static-to-Dynamic Model to improve Dynamic Facial Expression Recognition (DFER) in videos. BTN \cite{her2024batch} includes Multi-Level Attention (MLA) and Batch Transformer (BT) modules to address uncertainty and noisy data in Facial Expression Recognition (FER). ARBEx \cite{wasi2023arbex} incorporates learnable anchors and a multi-head self-attention mechanism in the embedding space to tackle class imbalance, bias, and uncertainty in expression learning tasks. 

Regarding continuous expression VA inference, CAGE \cite{wagner2024cage} is the current state-of-the-art framework. It uses a small-scale pre-trained version of the Multi-Axis Vision Transformer (MaxVIT) \cite{tu2022maxvit} along with a lightweight EfficientNet model \cite{tan2021efficientnetv2}. The core insight is to train both discrete category and VA estimation simultaneously, using the combined checkpoint to enhance VA inference.
Therefore, DDMFN and CAGE frameworks are employed as our 2D image-side backbones for fusion analysis.

\subsection*{Data Fusion} 

Intermediate fusion allows data fusion at different stages of model training by transforming input data into higher-level feature representations through multiple layers\cite{lahat2015multimodal}. It offers flexibility at different depths of fusion. In the context of deep learning with multimodal data, intermediate fusion involves merging different modal representations in a single hidden layer, enabling the model to learn joint representations. To improve performance, data dimensionality can be adjusted \cite{yi2015shared,ding2015robust}.

Late fusion involves independently processing data sources at the decision stage before fusing the results \cite{lahat2015multimodal}. This technique is inspired by the popularity of ensemble classifiers \cite{kuncheva2014combining}. When data sources differ significantly in sampling rate, data dimensionality, and measurement units, this technique is much simpler than early fusion methods. Since errors from multiple models are handled independently, they are uncorrelated, and late fusion typically yields better performance. 
Many researchers use late or decision-level fusion to address multimodal data problems \cite{simonyan2015two,wu2016deep,kahou2013combining}.

\section{Face Representation with 3D Morphable Models}

Although there are many different 3D face representation models, in this work, we are mainly considering 3D morphable models (3DMM), which have been widely used in widely used in facial geometry reconstruction tasks. 
These models regress the 3D representations of faces from 2D images, by projecting the input 2D image onto the previously established 3DMM space, to obtain shape, pose, expression, and detail coefficients. Nonlinear optimization techniques are then used to refine these parameters by minimizing a cost function. The pipeline demonstrating the relationship between face regression and 3DMM is shown in Figure \ref{fig:3dpipeline} 
Below, we briefly introduce the widely used FLAME model and the two latest FLAME-based regression tasks EMOCA and SMIRK. \medskip

\begin{figure*}[h]
    \centering
    \vspace*{-10pt}
    \includegraphics[width=0.90\linewidth]{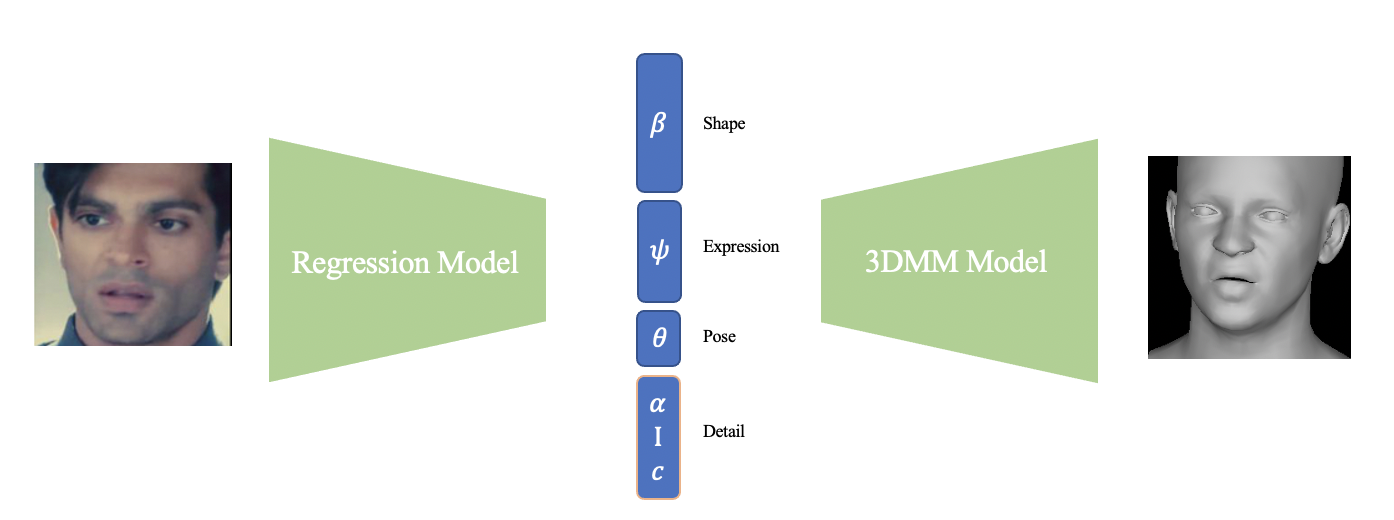}
    \caption{A standard pipeline for 3D facial geometry reconstruction from an image. Left: The regression model extracts disentangled 3D parameter representations from the images. Right: These parameters are utilized to reconstruct the 3D facial geometry using a 3D Morphable Model.}
    \label{fig:3dpipeline}
    \vspace*{-10pt}
\end{figure*}

\textbf{FLAME}\cite{FLAMESiggraphAsia2017} is a 3DMM used for synthesizing detailed and expressive 3D models of human heads. It accomplishes this by exploiting a linear shape space trained from a large dataset of 3D scan sequences (4D with time). The model uses Principal Component Analysis (PCA) to create a low-dimensional representation of facial shapes and this involves identifying the principal components (PCs) that capture the most variance in the facial shapes from the training data. Thus, any new face can be represented as a linear combination (the shape parameters) of these principal components. 
By adjusting the weights of these components, FLAME can generate a wide variety of facial shapes. To improve its fidelity, FLAME also includes pose-dependent corrective blendshapes and global expression blendshapes. The global blendshapes are predefined facial expressions that are added to the base facial shape. They capture various facial movements associated with emotions, such as smiling, frowning, or surprise. 

\begin{figure*}[!htbp]
    \centering
    \vspace*{-10pt}
    \includegraphics[width=0.80\linewidth]{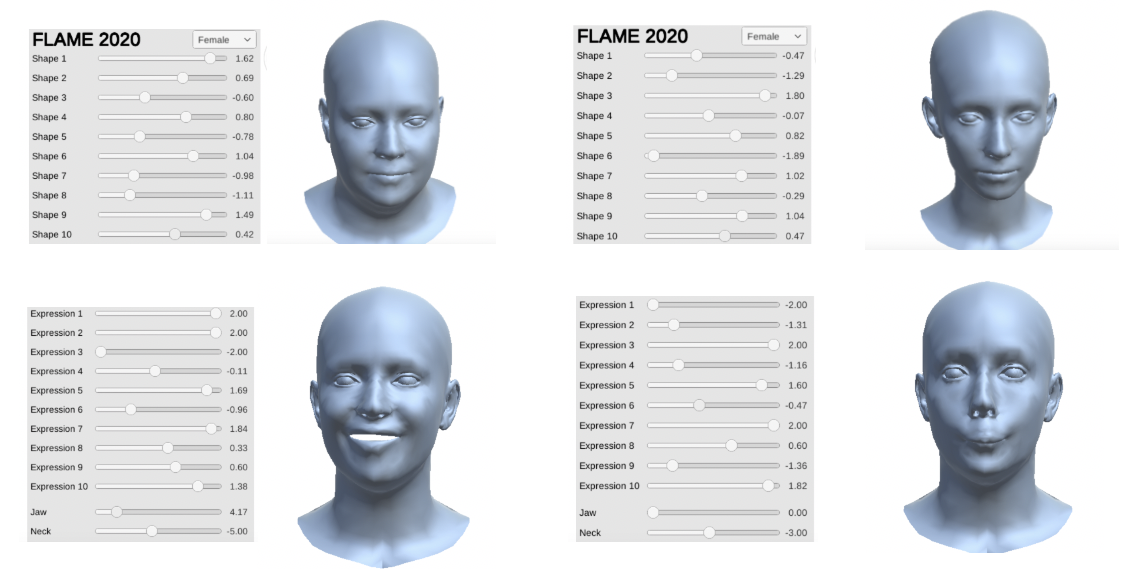}
    \caption{3D Representation Visualization}
    \label{fig:3Dfaces}
\end{figure*}

Figure \ref{fig:3Dfaces} shows a few examples of the resulting synthesized faces when various model parameters are altered. The images were created from the FLAME Model Viewer located on the authors' site\footnote{\href{https://flame.is.tue.mpg.de/}{https://flame.is.tue.mpg.de/}}. %

\textbf{EMOCA}\cite{danvevcek2022emoca} extends the previously trained face reconstruction framework, DECA \cite{feng2021learning}, which reconstructs a detailed 3D face model by learning the FLAME parameters from a single RGB image. 
The model uses a combination of landmark losses, photometric losses, and perceptual emotion consistency loss during training to ensure the reconstructed 3D faces accurately convey the emotional content of the input images.
The EMOCA model regresses a total of 334 parameters: 100 for shape, 50 for emotional expressions, 6 for pose, 100 for detail, 50 for texture, and others including pose-dependent and articulated components. \medskip

\textbf{SMIRK} \cite{smirk20243d} is another FLAME-based regression model that has the advantage of capturing any subtleties, extreme expressions, asymmetries, or rarely observed expressions that create slight deformations of the face shape.
SMIRK replaces the previous differentiable rendering approach in comparing generated 3D face representations with the original inputs. Given the rendered predicted mesh geometry and sparsely sampled pixels of the input image, this new neural rendering module focuses on local geometries to generate a face image more similar to the original, which can then be fed back to the reconstruction pipeline. 
The SMIRK model regresses to 358 standard parameters of which 300 are shape, 50 are expression and 6 are pose. Other additional parameters include camera parameters and those specific to the neural rendering process used in SMIRK.

\section{Fusion Architecture}

\subsection{3D Representation Classifier Architecture}

Though EMOCA and SMIRK report emotion recognition performance, their results are based on a manually cleaned dataset, which is difficult to reproduce. Despite the inherent limitations of the AffectNet dataset, for a fair comparison, we extracted all 3D features from AffectNet datasets to curate a 3D representation dataset and then trained the 3D representation classifier.

The classifiers in this study adopted a similar architecture to the emotion classification MLP proposed in \cite{toisoul2021estimation}, with minor adjustments. The classifier network architecture comprises an input layer, followed by four fully connected layers, each with an output dimension of 2048. Batch normalization and Leaky ReLU activation functions are applied in all fully connected layers. Dropout rates of 50\% and 40\% are applied to the first and second layers, respectively. The output layer is adapted based on the dataset: for the RAF-DB dataset, it has 7 output dimensions, while for the AffectNet dataset, it includes 8 classes plus additional outputs for valence and arousal levels.  
\begin{figure*}[h]
    \centering
    \vspace*{-10pt}
    \includegraphics[width=0.80\linewidth]{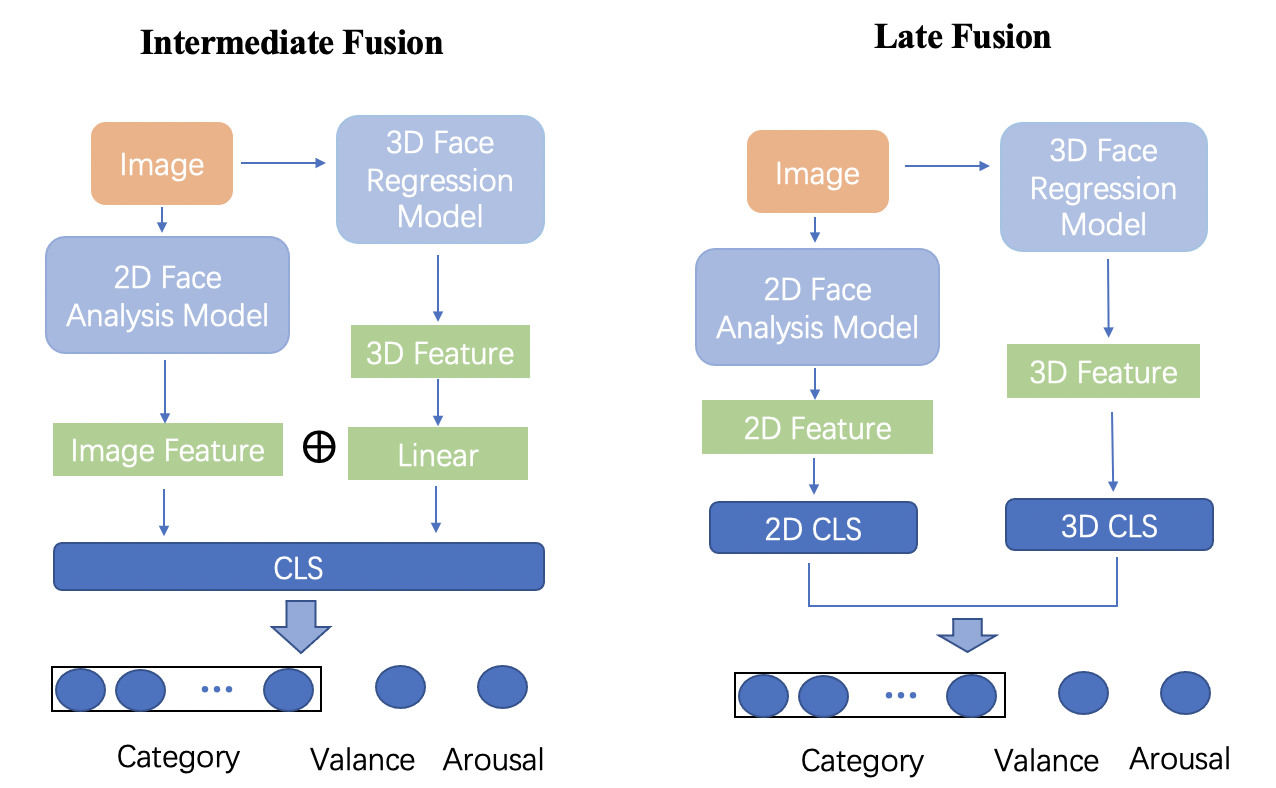}
    \caption{Overview of the 3D Representation Fusion Architecture}
    \label{fig:many}
\end{figure*}

\subsection{Loss Function }
\subsubsection{Discrete Expression Inference }
For the AffectNet 
dataset\cite{mollahosseini2017affectnet}, 
following the previous approach\cite{wagner2024cage, toisoul2021estimation} for emotion classification, we employed a combined loss function for this task.

The combined loss function used in this study integrates several components to handle both classification and regression tasks. The cross-entropy loss $L_{CE}$
 is used for the classification task, measuring the difference between the true and predicted class labels. For the regression tasks, the mean squared error $L_{MSE}$ is applied to minimize the squared differences between the predicted and true values of valence and arousal. Additionally, we incorporate the Pearson correlation coefficient  $L_{PCC}$, which assesses the linear correlation between the predicted and true valence and arousal values. Furthermore, the concordance correlation coefficient $L_{CCC}$ is used to measure the agreement between the predicted and true values, considering both precision and accuracy.

The final loss function is a weighted combination of these components as shown Eq: \ref{eq:loss1}:

\begin{equation}
\scriptsize
\begin{split}
    Loss = L_{CE} +\frac{\alpha}{\alpha+\beta+\gamma}\times L_{MSE} +\frac{\beta}{\alpha+\beta+\gamma}\times(1- L_{CCC}) + \frac{\gamma}{\alpha+\beta+\gamma}\times(1-L_{PCC})
\end{split}
\label{eq:loss1}
\end{equation}

where \(\alpha, \beta, \gamma\) are weighting factors for the different components of the loss function. The values of \(\alpha, \beta, \gamma\) are randomly sampled from a uniform distribution between 0 and 1 for each batch. 

\subsubsection{Continuous Expression Inference } Regarding the VA estimation on AffectNet, we follow CAGE \cite{wagner2024cage} and use a two-stage training approach and employ their best performance model, MaxVIT as our 2D image analysis model. In the first stage, we train with a combined loss $Loss_{combined}$. In the second stage, building on the first, we modify the output to valence and arousal to strengthen the VA supervised learning with $Loss_{va}$. 

\begin{equation}
\begin{split}
    Loss_{combined}=L_{weightedCE} +  w_{1}\cdot L_{MSE}
\end{split}
\label{eq:loss}
\end{equation}

\begin{equation}
\begin{split}
    Loss_{va}=L_{CCC} +  w_{2}\cdot L_{MSE}
\end{split}
\label{eq:loss}
\end{equation}

where the weighted cross-entropy $L_{weightedCE}$  is used to ease the issue of imbalanced distribution in the training set. This loss function assigns weights based on the frequencies of the expressions. 

\subsection{Intermediate Fusion Architecture}

In the intermediate fusion architecture shown in Fig. \ref{fig:many}, 
A facial image is processed by both a 2D face analysis model and a 3D face regression model. The 2D analysis model extracts high-dimensional features useful for recognizing facial expressions, while the 3D regression model regresses parameters that are linked to the facial structure.

Next, the 2D image features and the transformed 3D features are concatenated (or combined) into a comprehensive feature set. This feature set is then fed into a linear fusion architecture, leveraging both 2D and 3D information. The fused features are input into a classification layer (CLS), which is identical to the classifier architecture mentioned earlier. The CLS layer processes these features to predict various aspects of facial expressions (category, valence, or arousal). 

Different datasets necessitate different outputs and loss functions. The loss function and linear fusion architecture used here are consistent with the classifier architecture mentioned above.

\subsection{Late Fusion Architecture}
In the late fusion architecture shown in Fig. \ref{fig:many}, the input image is processed by the 2D face analysis model. This model extracts 2D features from the image and inputs these 2D features into the 2D classifier, which outputs emotion category, valence, and arousal results based on the requirements of different datasets. Meanwhile, the input image is also processed by the 3D face regression model. This model extracts 3D features from the image, and these 3D features are then input into a pre-trained classifier, which outputs emotion category, valence, and arousal results. 
This architecture maximizes the preservation of their respective independent inference capabilities.

In the late fusion step, the outputs of the 2D classifier and the 3D classifier are combined. Here, we use simple fusion methods such as max, mean, and weighted to integrate the information from both 2D and 3D analyses, generating more accurate and robust emotion recognition results.

\section{Experiments}

\subsubsection{Training Setting and Evaluation Metrics}

\begin{table*}[h]
\small
\caption{ \textbf{Hyperparameters for Classifier Training.}}
    \vspace{-0.2 cm}
    \centering
    \begin{tabular}{|l|c|}
        \hline
        \textbf{Hyperparameter} & \textbf{Value} \\
        \hline
        Batch Size & 64 \\
        \hline
        Weight Decay & 1e-5 \\
        \hline
        Maximum Epochs & 100 \\
        \hline
        Early Stopping Patience & 3 \\
        \hline
        Learning Rate Scheduler & CyclicLR \\
        \hline
        Base Learning Rate & 1e-6 \\
        \hline
        Maximum Learning Rate & 1e-4 \\
        \hline
        Step Size & \texttt{len(train\_loader)//2} \\
        \hline
        Scheduler Mode & 'triangular' \\
        \hline
        Cyclical Momentum & No \\
        \hline
    \end{tabular}
    
    \label{tab:hyperparameters}
\end{table*}

The hyperparameters for the training architecture are summarized in Tab. \ref{tab:hyperparameters}. These settings ensured the stability and efficiency of the model training process.

For the emotion discrete classification task, we use state-of-the-art binary classification metrics \cite{zhang2023dual, wagner2024cage}, including Accuracy, F1 score, Precision, and Recall. We follow established practices\footnote{\href{https://github.com/wagner-niklas/CAGE_expression_inference/blob/main/models/evaluation.py/}{Code reference for evaluation (see lines 90-92).}} to compute these metrics. For a comprehensive evaluation, we report both weighted and macro averages for the unbalanced test dataset (RAF-DB). For the balanced test dataset (AffectNet), where both metrics are identical, we report results once.

For the emotion continuous regression task, our evaluation metrics are Mean Squared Error (MSE), Mean Absolute Error (MAE), Root Mean Squared Error (RMSE), and Concordance Correlation Coefficient (CCC).

\vspace{-0.5 cm}

\begin{table*}[htbp]
\small
\caption{ \textbf{Classification Comparison of EMOCA and SMRIK 3D Representations only (no fusion) on AffectNet Dataset.}}
\vspace{-0.3 cm}
\begin{center}
\begin{tabular}{|c|c|c|c|c|}
\hline
\textbf{ 3D Classifier } & \textbf{Accuracy} $\uparrow$ & \textbf{F1} $\uparrow$  &  \textbf{Precision} $\uparrow$ & \textbf{Recall} $\uparrow$ \\
\hline
 $CLS_{Smirk3D-short}$   & 0.5461 & 0.5459 & 0.5477 & 0.5461 \\
\hline
 $CLS_{Smirk3D-full}$   & 0.5546 & 0.5547 & 0.5569 & 0.5546 \\
\hline
 $CLS_{Emoca3D-short}$ &  \textbf{0.5723}  & \textbf{0.5726}  & \textbf{0.5758}  & \textbf{0.5723}  \\
\hline
 $CLS_{Emoca3D-full}$  & 0.5703 & 0.5704 & 0.5768 & 0.5703 \\
\hline
\end{tabular}
\label{AF_CLS_acc}
\end{center}
\end{table*}
\FloatBarrier 
\vspace{-1.0 cm}

\begin{table}[htbp]
\small
\caption{ \textbf{Classification Comparison of EMOCA and SMRIK 3D Representations only (no fusion) on RAF-DB Dataset.} Due to the unbalanced test dataset, we report both weighted and macro average metrics for a comprehensive evaluation. \textbf{Acc} stands for Accuracy, \textbf{F1} for F1 score, \textbf{P} for Precision, and \textbf{R} for Recall.}
\vspace{-0.3 cm}
\begin{center}
\begin{tabular}{|c|c|c|c|c|c|c|c|}
\hline
\multirow{2}{*}{\textbf{3D Classifier}} & \multirow{2}{*}{\textbf{Acc} $\uparrow$} & \multicolumn{3}{c|}{\textbf{Weighted Avg}} & \multicolumn{3}{c|}{\textbf{Macro Avg}} \\
\cline{3-8} 
 &  & \textbf{F1} $\uparrow$ & \textbf{P} $\uparrow$ & \textbf{R} $\uparrow$ & \textbf{F1} $\uparrow$ & \textbf{P} $\uparrow$ & \textbf{R} $\uparrow$ \\
\hline
 $CLS_{Smirk3D-short}$   & 0.7378  & 0.7418  & 0.7475  & 0.7418 & 0.6421  & 0.6386  & 0.6482 \\
\hline
 $CLS_{Smirk3D-full}$   & 0.7557  & 0.7584 &  0.7631& 0.7557  & 0.6585 &  0.6568  & 0.6627 \\
\hline
 $CLS_{Emoca3D-short}$ &  0.7862  & 0.7873  & 0.7895  & 0.7862 & 0.6965  & 0.6908  & 0.7037 \\
\hline
 $CLS_{Emoca3D-full}$  & \textbf{0.7927}  & \textbf{0.7946}  & \textbf{0.7985} & \textbf{0.7927} & \textbf{0.7073}  & \textbf{0.7043} & \textbf{0.7118}  \\
\hline
\end{tabular}
\label{RAF_CLS_ACC}
\end{center}
\vspace{-0.5 cm}
\end{table}

\subsection{ 3D Representation Classification Performance}

In 3D representations, the parameters most relevant to expressions are expression, pose, and shape. To analyze the information gained from other parameters on FEI tasks, we divided them into two groups: the short group (expression, pose, shape) and the full group (all parameters). Thus, $Smirk3D_{short}$ has 353 dimensions, $Smirk3D_{full}$ has 358 dimensions, $Emoca3D_{short}$ has 156 dimensions, and $Emoca3D_{full}$ has 334 dimensions.

As shown in Tab. \ref{AF_CLS_acc} and Tab. \ref{RAF_CLS_ACC}, $Emoca3D$ outperforms $Smirk3D$ in discrete emotion classification tasks. $Emoca3D_{short}$ achieves the best performance on AffectNet, improving by 2.26\% in accuracy compared to $Smirk3D_{short}$, while $Smirk3D_{full}$ achieves the best performance on RAF-DB, improving by 3.7\% in accuracy compared to $Smirk3D_{full}$. Experiments show that EMOCA preserves more emotional information and is better suited for emotion reasoning tasks.

\subsection{ 3D Fusion in Discrete Facial Expression Inference}

In the experiments of discrete expression inference on the AffectNet dataset as shown in Tab. \ref{affect_fusion_evaluation}, our fusion method achieved the best performance. 
Although we did not replicate its best performance as reported (65.04\%) on AffectNet 8, the weighted late fusion still achieved an improvement in all metrics. Here, the late fusion weight is 0.2 for the 3D component. The weighted fusion strategy improved the accuracy by 0.55\%, the F1 score by 0.58\%, the precision by 0.26\%, and the recall by 0.55\%.

\begin{table}[h] 
\small
\caption{ \textbf{Classification Comparison of Different Fusion Architectures on AffectNet Dataset.}}
\vspace{-0.3 cm}
\label{affect_fusion_evaluation}
\begin{center}
\begin{tabular}{|c|c|c|c|c|}
\hline
\textbf{ Framework } & \textbf{Accuracy} $\uparrow$ & \textbf{F1} $\uparrow$  &  \textbf{Precision} $\uparrow$ & \textbf{Recall} $\uparrow$ \\
\hline
DDAMFN (our reproduction)  & 0.6324 & 0.6323  & 0.6353 & 0.6324\\
\hline
\textbf{Intermediate Fusion} & &  & & \\
\hline
 $F_{2D}$ + $F_{Smirk3D}$  & 0.6117 & 0.6098 & 0.6128 & 0.6117 \\
\hline
 $F_{2D}$ + $F_{Emoca3D}$ & 0.6234 & 0.6232  & 0.6276  & 0.6234 \\
\hline
\textbf{Late Fusion} & &  & &\\
\hline
 Max with  $CLS_{Smirk3D}$ & 0.6267 & 0.6260  & 0.6273 & 0.6267 \\
 Max with $CLS_{Emoca3D}$ & 0.6294  & 0.6292 & 0.6306 & 0.6294\\
\hline
 Mean with $CLS_{Smirk3D}$& 0.6262 & 0.6266  & 0.6315  & 0.6262  \\
 Mean with $CLS_{Emoca3D}$& 0.6289 &  0.6295 & 0.6338 & 0.6289 \\
\hline
Weighted with $CLS_{Smirk3D}$ & 0.6364  & 0.6367  & 0.6408  & 0.6364  \\
Weighted with $CLS_{Emoca3D}$ & \textbf{0.6379}  & \textbf{0.6381}  & \textbf{0.6379}  & \textbf{0.6379} \\
\hline
\end{tabular}
\end{center}
\vspace{-0.5 cm}
\end{table}

\begin{table}[htbp] 
\small
\caption{ \textbf{Classification Comparison of Different Fusion Architectures on RAF-DB Dataset.} Due to the unbalanced test dataset, we report both weighted and macro average metrics for a comprehensive evaluation. \textbf{Acc} stands for Accuracy, \textbf{F1} for F1 score, \textbf{P} for Precision, and \textbf{R} for Recall.}
\vspace{-0.3 cm}
\begin{center}
\begin{tabular}{|c|c|c|c|c|c|c|c|}
\hline
\multirow{2}{*}{\textbf{Framework}} & \multirow{2}{*}{\textbf{Acc} $\uparrow$} & \multicolumn{3}{c|}{\textbf{Weighted Avg}} & \multicolumn{3}{c|}{\textbf{Macro Avg}} \\
\cline{3-8}
 &  & \textbf{F1} $\uparrow$ & \textbf{P} $\uparrow$ & \textbf{R} $\uparrow$ & \textbf{F1} $\uparrow$ & \textbf{P} $\uparrow$ & \textbf{R} $\uparrow$ \\
\hline
DDAMFN (our reproduction)  & 0.9016 & 0.9013  & 0.9022 & 0.9016 & 0.8554  & 0.8686 & 0.8451 \\
\hline
\textbf{ Intermediate Fusion} & &  & & & & &\\
\hline
 $F_{2D}$ + $F_{Smirk3D}$  & 0.9006 & 0.9007 & 0.9018 & 0.9006 & 0.8489 & 0.8561 & 0.8435 \\
\hline
 $F_{2D}$ + $F_{Emoca3D}$ & 0.8996 & 0.8990  & 0.8989 & 0.8996 & 0.8501 & 0.8559 & 0.8453  \\
\hline
\textbf{Late Fusion} & & & & & & &\\
\hline
 Max with $CLS_{Smirk3D}$ & 0.8989 & 0.8984  & 0.8989 & 0.8989 & 0.8527 & 0.8656 & 0.8426 \\
 Max with $CLS_{Emoca3D}$ &0.8941 & 0.8944 & 0.9021 & 0.8941 & 0.8462 & 0.8643 & 0.8485 \\
\hline
 Mean with $CLS_{Smirk3D}$& 0.9030 & 0.9024 & 0.9041& 0.9030 &  0.8561 &  0.8829 & 0.8361 \\
 Mean with $CLS_{Emoca3D}$& 0.9130 & 0.9135 & 0.9178 & 0.9130 &  0.8413 &  0.8414 & 0.8521 \\
\hline
Weighted with $CLS_{Smirk3D}$ & 0.9106  & 0.9099  & 0.9110  & 0.9106 &  0.8689 &  0.8914 & 0.8516 \\
Weighted with $CLS_{Emoca3D}$ & \textbf{0.9400}  & \textbf{0.9393}  & \textbf{0.9397}  & \textbf{0.9400} & \textbf{0.8958}  & \textbf{0.9090}  & \textbf{0.8860}\\
\hline
\end{tabular}
\label{raf_DB_fusion_evaluation}
\end{center}
\vspace{-0.5 cm}
\end{table}

The best performance was achieved with the weighted Emoca3D late fusion strategy, reaching the highest accuracy (94.00\%), F1 score (93.93\%), precision (93.97\%), and recall (94.00\%). Compared to our reproduced DDAMFN model,
the weighted fusion strategy improved the accuracy by 3.84\%, the F1 score by 3.80,\% the precision by 3.75\%, and the recall by 3.84\%. This result has made our model overpass the state-of-the-art performance on the RAF-DB dataset.

We believe the reason late fusion outperforms intermediate fusion is that the results from different models remain more independent, thereby maximizing the retention of each model's respective advantages.
However, the features provided by intermediate fusion are redundant compared to the original features. 
During training, we observed early stopping occurring within the first 8 epochs, earlier than the baseline model's 12-15 epochs. We guess this is because intermediate fusion introduces excessive redundant information, which strongly correlates with features extracted from images in high-dimensional space, leading to reduced model generalization.

\vspace{-0.5 cm}
\begin{table}[h]
\small
\caption{ \textbf{Comparison with Previous SOTA models for Discrete FEI on RAF-DB Dataset.}}
\vspace{-0.3 cm}
\begin{center}
\begin{tabular}{|c|c|c|}
\hline
\textbf{ Method } & \textbf{Accuracy [\%]} & \textbf{ Date [mm-yy] }\\
\hline
FMAE \cite{ning2024representation}  & 93.09 & 07-2024\\
\hline
S2D \cite{chen2023static}  & 92.57 & 12-2023\\
\hline
BTN \cite{her2024batch}  & 92.54 & 07-2024\\
\hline
ARBEx \cite{wasi2023arbex} & 92.37 & 05-2023\\
\hline
DDAMFN \cite{zhang2023dual} & 92.34 & 07-2023\\
\hline
Ours  & \textbf{94.00} & 07-2024\\
\hline
\end{tabular}
\label{table_raf_benchmark}
\end{center}
\vspace{-0.6 cm}
\end{table}

The results shown in Tab.\ref{table_raf_benchmark} indicate that a simple weighted late fusion strategy can significantly enhance the performance of our reproduced DDAMFN model (originally ranked 5th), achieving new state-of-the-art performance. Note, that the TOP1 FMAE model pre-trained on the large dataset, while our method just trained with a single AffectNet dataset.

\begin{table}[h]
\small
\caption{ \textbf{Continuous VA Results from Different Fusion Architectures on AffectNet Dataset.}}
\vspace{-0.3 cm}
\begin{center}
\begin{tabular}{|c|c|c|c|c|}
\hline
\textbf{ Framework } & \textbf{MSE} $\downarrow$ & \textbf{MAE} $\downarrow$  &  \textbf{RMSE} $\downarrow$& \textbf{CCC} $\uparrow$ \\
\hline
$CAGE_{va}$  (Our reproduction)  & 0.1044 & 0.2377  & 0.3230 & 0.7814\\
\hline
\textbf{3D Representation} &  & &   & \\
\hline
 $Regresser_{Emoca3D}$ & 0.1061 & 0.2483  & 0.3257 & 0.7637 \\
\hline
\textbf{Feature Fusion} & &  & & \\
\hline
 $F_{2D}+F_{Emoca3D}$  & 0.1061 & 0.2398  & 0.3257 & 0.7749 \\
\hline
\textbf{Late Fusion} & &  & &\\
\hline
 Max with $Regressor_{Emoca3D}$  & 0.1052 & 0.2419 & 0.3243 & 0.7727 \\
\hline
 Min with $Regressor_{Emoca3D}$ &0.1053 & 0.2441 & 0.3245 & 0.7726  \\
\hline
 Mean with $Regresser_{Emoca3D}$ &  \textbf{0.0956} & 0.2325  & \textbf{0.3092} & 0.7891 \\
 \hline
Weighted with $Regresser_{Emoca3D}$ & 0.0958  & \textbf{0.2316}  & 0.3095 & \textbf{0.7901}\\
\hline
\end{tabular}
\label{table_va}
\end{center}
\end{table}

\begin{table}[h]
\small
\caption{ \textbf{Benchmark Comparison for VA Inference on AffectNet Dataset.}}
\vspace{-0.3 cm}
\label{table_evaluation}
\begin{center}
\begin{tabular}{|c|c|c|c|c|c|}
\hline
\textbf{ Framework } & \textbf{RMSE\(_{val}\)}$\downarrow$ &  \textbf{RMSE\(_{aro}\)}$\downarrow$  & \textbf{CCC\(_{val}\)}  $\uparrow$&  \textbf{CCC\(_{aro}\)} $\uparrow$  & \textbf{Date[mm-yy]} \\
\hline
VGG-G \cite{bulat2022pre}  & 0.356 & 0.326  & 0.710 & 0.629 & 03-2021\\
\hline
CAGE \cite{wagner2024cage}  & 0.331 & 0.305  & 0.716 & 0.642 & 04-2024\\
\hline
Ours   & \textbf{0.323}  & \textbf{0.294}  & \textbf{0.724} &  \textbf{0.650} & 07-2024\\
\hline
\end{tabular}
\label{va_ben}
\end{center}
\vspace{-0.5 cm}
\end{table}

\subsection{ 3D Fusion in Continuous Facial Expression Inference}

The experimental results on AffectNet, as shown in Table \ref{table_va}, indicate that our late fusion and mean fusion strategies effectively improved performance. Surprisingly, the performance of the 3D representation alone is already very close to that of our reproduced CAGE.
Our mean fusion achieved an MSE of 0.0956 and an RMSE of 0.392, while our weighted fusion achieved an MAE of 0.2316 and a CCC of 0.7901. The weight for the 3D representation was set at 0.4. These results highlight the importance of 3D representation to continuous FEI tasks.
Compared to our reproduced CAGE model, our mean late fusion improved performance by 8.43\% in MSE, 2.19\% in MAE, 4.27\% in RMSE, and 0.99\% in CCC. Our weighted late fusion increased performance by 8.24\% in MSE, 2.57\% in MAE, 4.18\% in RMSE, and 1.11\% in CCC.

Tab.\ref{va_ben} shows that our late fusion method has surpassed the current state-of-the-art methods. Compared with the CAGE model, our valence RMSE increased by 2.42\% ,  arousal RMSE increased by 3.61\%, valence CCC increased by 1.12\%, and arousal CCV increased by 1.25\%. The overall results indicate that the late fusion of 3D representations can effectively enhance continuous expression inference tasks.

The experimental results on the AffectNet as shown in Tab.
\ref{table_va}, our late fusion and mean fusion both greatly improved the performance. Here the weight for 3D representation is 0.4.
Our mean fusion achieved 0.0956 in MSE and 0.392 in RMSE, meanwhile, our weighted fusion achieved 0,2316 in MAE and 0.7901 in CCC. 
This highlights the 3D representation's contribution to continuous FEI tasks.
Our mean fusion achieved an MSE of 0.0956 and an RMSE of 0.392, while our weighted fusion achieved an MAE of 0.2316 and a CCC of 0.7901. 
Compared to our reproduced CAGE model, our mean late fusion performance improved by 8.43\% in MSE, 2.19\% in MAE, 4.27\% in RMSE, and 0.99\% in CCC. Our weighted late fusion performance increased by 8.24\% in MSE, 2.57\% in MAE, 4.18\% in RMSE, and 1.11\% in CCC.

\section{ Conclusions}

As 3D face reconstruction aligns more closely with reality, its integration and analysis can be beneficial in many FEI tasks. Our research evaluates the performance of existing 3D face representations, introduces two fusion architectures, and demonstrates the efficiency of late fusion through extensive experiments.
The experimental results show that our proposed method outperforms the state-of-the-art in AffectNet VA estimation and RAF-DB classification tasks. These findings offer valuable insights into the application of 3D representations for emotion inference.

Moving forward, our future work will focus on several key areas. First, we will delve deeper into the analysis of 3D representations within the realm of micro-expressions, aiming to capture even the most subtle emotional cues. Second, we will investigate how 3D reconstructions can enhance emotion inference in scenarios involving human backgrounds and human-human interactions. By continuing to refine and expand upon these methodologies, we aim to contribute to the development of more nuanced and accurate emotion inference models.

\section*{Acknowledgments}
This work is based upon work supported under the National AI Research Institutes program by the National Science Foundation (NSF) and the Institute of Education Sciences (IES), U.S. Department of Education, through Award \# 2229873 and NSF Award \# 2223507. Any opinions, findings and conclusions or recommendations expressed in this material are those of the author(s) and do not necessarily reflect the views of the NSF, the IES, or the U.S. Department of Education.

\newpage
\bibliographystyle{splncs04}
\bibliography{main}

\end{document}